\documentclass[11pt]{article}
\usepackage{coling2020}
\usepackage{times}
\usepackage{url}
\usepackage{latexsym}

\usepackage{amsthm}

\usepackage{caption}
\usepackage{subcaption} 
\usepackage{amsmath,amssymb,amsfonts}
\usepackage{algorithmic}
\usepackage{graphicx}
\usepackage{textcomp}
\usepackage{xcolor}
\usepackage{float}

\usepackage{multirow}

\newcommand{\fig}{Figure~}
\newcommand{\tab}{Table~}

\newcommand{\B}{\mathcal{B}}
\newcommand{\E}{\mathbb{E}}

\title{Hybrid Supervised Reinforced Model for Dialogue Systems}

\author{Carlos Miranda \\
  Worldline / Seclin, France \\
  INSA de Rouen / Rouen, France \\
    {\tt carlos.miranda\_lopez@insa-rouen.fr} \\\And
  Yacine Kessaci \\
  Worldline / Seclin, France \\
  {\tt yacine.kessaci@worldline.com} \\}
	
\date{}

\begin{document}
\maketitle
\begin{abstract}
This paper presents a recurrent hybrid model and training procedure for task-oriented
dialogue systems based on Deep Recurrent Q-Networks (DRQN).  The model copes with both tasks required for Dialogue Management: State Tracking and
Decision Making. It is based on modeling Human-Machine interaction into a latent representation embedding an interaction context to guide the discussion. The model achieves greater performance, learning speed and
robustness than a non-recurrent baseline. Moreover, results
allow interpreting and validating the policy evolution and the latent
representations information-wise.
\end{abstract}

%

\section{Introduction}\label{sec:introduction}

\begin{figure}[hbtp]
    \centering
    \includegraphics[width=0.35\textwidth]{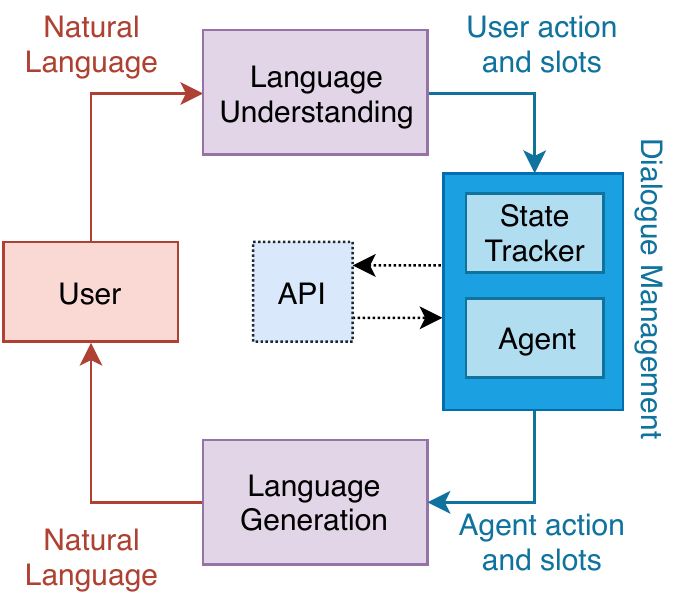}
    \caption{\textbf{Dialogue System pipeline: a language understanding (LU) unit yields
a representation for a Dialogue Management (DM) unit, composed of a State Tracker (ST) and an Agent. The latter decides on an
action, which is given to the Language Generation (LG) unit to translate into
Natural Language (NL).}}
    \label{fig:ds_model}
\end{figure}

Conversational agents - chatbots, virtual
assistants and voicecontrol interfaces - are becoming omnipresent in modern
society. Applications include connected
devices, internal support, and even sales automation. There is a
growing demand on the performance of these goal-based systems: tasks become
more complex so there is a need for new and better techniques in this domain.
That is why traditional rule-based approaches in dialogue systems are being
superseded by data-driven learning systems. Indeed, they provide more
robustness, as well as richer dialogue state representations and
non-stationary interaction strategies, which account for more human-like
behaviour. One such system can be modeled as in \fig \ref{fig:ds_model}.

Data-driven learning systems make use of Machine Learning (ML) to deal with
both Human-Machine interfaces (speech and text recognition) and Dialogue
Management. Nowadays, the trend is to use Supervised Learning (SL) to train
some, if not all, of these systems' components. Specifically for
Dialogue Management, both rule-based and supervised learning agents remain
the most used approaches \cite{chen17survey}. However, using a set of rules
for dialogue management leads to agents that cannot deal with undefined
situations with a stationary response selection strategy. As for SL agents,
they learn from a labeled training set, so they can only reproduce dialogue
strategies that are present in the said data set which does not represent 
necessarily the optimal strategies. That is why some novel training strategies 
for dialogue management opt for Reinforcement Learning agents
\cite{cuayahuitl15strategicdm,henderson13wordbaseddst}, which can find
optimal dialogue policies in Partially Observable Markov Decision
Process-based (POMDP) scenarios.

Moreover, some works introduced end-to-end training of such data driven dialogue systems
\cite{li17tcbot}, where the system is trained as a whole, or at least some
parts are trained jointly. In our case, we advocate for a joint training of
the DM -- the state tracker (ST) and the agent. In fact, recent advances in
Recurrent Reinforcement Learning (RRL) have shown good results in POMDP-based
scenarios \cite{hausknecht15drqn}, but in significantly different
environments.

In a Dialogue System setting, an agent does not have full
access to the user's goal, and has to deal with noise coming from the NL
units, so we can model it as a PODMP \cite{williams07pomdpsds}. Thus, our
work aims at proposing both a new architecture and training procedure in order to apply RRL
to efficiently learn a state tracking--action selection strategy. By using recurrent models we intend to build richer representations of states than those provided by classic state trackers. In addition, by jointly training our model we aim at establishing a more complex but direct interface between these two components. We have also explored how to use the intermediate representation of our
recurrent models to assess learning.

Our paper is structured as follows. In section \ref{sec:dm_with_rl}, we give a brief introduction of reinforcement learning for chatbots by defining the two main components of Dialogue Management. In Section \ref{sec:environment}, we give some elements that differentiate our environment to what we call \textit{arcade environments}, used in literature for state-of-the-art models. In Section \ref{sec:proposed_approaches}, we explain the model that we propose. Finally, Section \ref{sec:experiments} presents our results in a dialogue system testbed called TC-Bot and we discuss interpretability of the learned model. To the best of our knowledge, our recurrent joint Dialogue Management training is the first to achieve significant results in this setting and yielding insights of what has been learned.
%
%


\section{Dialogue Management with Reinforcement Learning}\label{sec:dm_with_rl}

Belief tracking is one of two tasks making up dialogue management in RL Dialogue
Systems. In POMDP-based Reinforcement Learning, the agent cannot perceive the
full states of the environment, it gets an observation, i.e. a partial
representation of the state. A \textit{belief tracker}, also called state
tracker (ST), takes a representation of the latest user utterance and
synthesizes a \textit{belief state} that sums up the dialogue until that
point. Depending on the representation provided by the NLU unit, state
trackers inputs can take multiple shapes. In our case, an observation is a
vector $o_t$ composed of binary vectors, where each dimension accounts for an
information slot (0 means empty, 1 means filled), one-hot-encoded (OHE)
vectors and float vectors (see \tab \ref{fig:observation}). Using this
model, the most simple change over the state tracker will require only to perform a binary OR operation in
slot-filled vector representations, so that the agent knows which slots have
already been filled or have been filled during the latest transition.

One can model the state tracker as a mapping $\phi$ between a pair
\textit{(previous belief state $h_{t-1}$, current observation $o_t$)} into
some representation on a space $\mathcal{B}$, i.e.

\begin{equation}\label{eq:belief_tracker} h_t = \phi(o_t, h_{t-1}), \quad
h_t \in \mathcal{B} \end{equation}

Even if it seems natural to some extent to put the state tracker in the
dialogue management unit, some models include it within the NLU unit --
intuiting the understanding can benefit from dialogue context and vice versa.
Thus, NLU and state tracker can actually perform a joint slot-filling task
\cite{rastogi2018multi} to yield a binary vector representation of the current belief state.
However, this training does not allow the state tracker and the agent to
exchange error feedbacks and leads the belief state to have some arbitrary representations.

That is why we advocate for a joint training of the state tracker and the RL
agent. In this way, a communication between both
entities is established, and the representation learned by the state tracker is directly
influenced by the learning of the agent: if the agent does bad, it is in part due to
a bad representation of the state. Moreover, this allows a
more flexible representation of the belief state, where the state tracker
is not constrained to a slot-filling task but can provide richer
representations to the agent. In our work, we use a Recurrent Neural Network (RNN) to approximate $\phi$.

Policy optimization is the second task in dialogue management. We want our
agent to learn an optimal policy $\pi^*$ to complete a task. A policy $\pi$
is a probability over actions given the current belief state, $\pi : \B
\times A \rightarrow [0; 1]$,where $A$ is the set of possible actions at
current time $t$. In order to learn such a policy, we can apply any
reinforcement learning methods.

Value-based RL methods approximate a state value function $V^\pi$ or a
state-action value function $Q^\pi$. These functions give a \textit{utility}
score for a given state or state-action pair by computing the expected value
of some notion of reward, represented by a random variable $R_t$, i.e.

\begin{align}\label{eq:value_functions}
    V^\pi(s_0)&= \E\left[R_t \vert s_0, \pi \right] \\
    Q^\pi(s_0, a_0)&= \E\left[ R_t \vert s_0, a_0, \pi\right] 
\end{align}

These value functions can then be used to obtain a policy. For example, by
choosing the action with the maximum $Q$-value with probability $1 -
\epsilon$ and a random action with probability $\epsilon$, we obtain the
$\epsilon$-\textit{greedy} policy.

Policy-based RL methods parameterize the policy $\pi_\theta$ with a vector
parameter $\theta$ and apply discrete or continuous optimization methods on
this function.

Finally, Actor-Critic methods apply both of these approaches so as to compute
a value function and approximate a policy \cite{barto83actorcritic}. 

Note that the RL algorithm addressed in this work - Deep Q-Network (DQN) - uses a value-based RL method with a state-action value function.
\section{Chatbot environment}\label{sec:environment}

All the aforementioned techniques, and many variations, have been successfully used to
train reinforcement learning agents \cite{mnih16a3c,hessel17rainbow,espeholt18impala}. In practice, one faces recurring problems when training an
RL agent, like the access to data and evaluation homogeneity. In fact, unlike
in supervised learning, data cannot be collected and reused, at least not
with the same easiness. Moreover, most literature RL results are achieved in
video-game simulators, e.g. the Arcade Learning Environment (ALE)
\cite{bellemare13arcade} or the DeepMind Lab (DMLab) \cite{beattie16dmlab}.
In our case, the environment is quite different as our agent will interact
with humans. We believe one of the most important differences is the length
of the episodes, as in most cases we want dialogues to be the shortest
possible, but it also depends on the user. Then, in practice, some techniques
may not scale down well enough, hyper-parameter values can be completely
different to state-of-the-art models, and comparison can be quite difficult.

In order to assess the data accessibility and evaluation homogeneity
problems, Li et al. proposed TC-Bot \cite{li17tcbot}, an end-to-end
training framework allowing to train a model on high-level representations of
user utterances. We used this framework as both a testbed and building block
for our approach. The framework is based on Agenda-Based User Simulator (ABUS) 
\cite{schatzmann07abus}. It allowed us to focus on the dialogue
management unit and provided a baseline to compare to. Moreover, recent work
\cite{kreyssig18nus} pointed at some problems with ABUS. In fact, as
originally conceived, the simulator represents a rather stationary
environment and does not account for noise coming from other components on
the loop. That is why Li et al. \cite{li17tcbot} introduced an Error Model
Controller (EMC) that allows to introduce noise at different levels: intent
level -- intent is not correctly recognized -- and slot level -- slot or
values are not correctly recognized.
\section{Proposed approach}\label{sec:proposed_approaches}

We examined the use of multiple architectures and training techniques. As a
matter of fact, our first model did not show competing performance, so we
expose which problems we faced and how we solved them. In this section, we
provide insights on the evolution of our model and training procedure.

As a common ground, all of the models have a recurrent layer and take
sequences of observations as input. Each observation is composed of 9
subvectors (see \tab \ref{fig:observation}). In addition, the models use the
Long-Short Term Memory (LSTM) \cite{hochreiter97lstm} variant, but similar
performance was reached with Gated Recurrent Unit (GRU) \cite{cho14gru}.
While our main objective is to jointly train ST and agent, we do not
completely remove the handcrafted ST from the framework. In fact, it is a
quite simple but effective ST. Thus, we used the part that allows to have a friendlier
representation of the current state for other components. We indirectly get
this handcrafted ST information through the database results subvectors.
Nonetheless, we could easily extend our model to produce slot-filled
representations, but we actually wanted to produce richer representations, not
restricted to the deduced one. Zhao et al. actually proposed a purely RL model
\cite{zhao16e2erl} where the agent must do both: keep track of the belief
state in the form of a binary vector and select actions. More details are given in Section \ref{sec:related_works}.

\begin{table*}
    \centering
    \footnotesize
    \begin{tabular}{|c|c|c|c|c|c|c|c|c|c|c|c|}
        \hline
        \multirow{2}{*}{Meaning} & \multicolumn{3}{c|}{User}                                                                                                                                                 & \multicolumn{3}{c|}{Agent}                                                                                                                                                & \multicolumn{5}{c|}{Dialogue}                                                                                                                                                                           \\ \cline{2-12} 
                                 & \begin{tabular}[c]{@{}c@{}}Dialogue\\ Act\end{tabular} & \begin{tabular}[c]{@{}c@{}}Inform\\ Slots\end{tabular} & \begin{tabular}[c]{@{}c@{}}Request\\ Slots\end{tabular} & \begin{tabular}[c]{@{}c@{}}Dialogue\\ Act\end{tabular} & \begin{tabular}[c]{@{}c@{}}Inform\\ Slots\end{tabular} & \begin{tabular}[c]{@{}c@{}}Request\\ Slots\end{tabular} & \textbf{\begin{tabular}[c]{@{}c@{}}Current\\ slots\end{tabular}} & Turn  & \textbf{Turn} & \begin{tabular}[c]{@{}c@{}}DB \\ results\end{tabular} & \begin{tabular}[c]{@{}c@{}}DB\\ results\end{tabular} \\ \hline
        Type                     & OHE                                                    & Binary                                                 & Binary                                                  & OHE                                                    & Binary                                                 & Binary                                                  & Binary                                                           & Float & OHE           & Binary                                                & Float                                                \\ \hline

        Notation                     & $u^{act}$                                                    & $u^{is}$                                                 & $u^{rs}$                                                  & $a^{act}$                                                    & $a^{is}$                                                 & $a^{rs}$                                                  & -                                                           & $t$ & -           & $kb^{bin}$                                                & $kb^{cnt}$                                                \\ \hline
        \end{tabular}
    \caption{\textbf{Components of an observation -- an observation is made up by
    concatenating these vectors. Database (DB) results are binary
    (constraint satisfied or not) and float counts (how many rows satisfy each constraint) of results
    matching all the current constraints perceived by the agent. Vectors in \textbf{bold} were dropped from the initial observations given by TC-Bot.}}
    \label{fig:observation}
\end{table*}

 \begin{minipage}{\linewidth}
      \centering
      \begin{minipage}{0.4\linewidth}
          \begin{figure}[H]
						\includegraphics[width=\textwidth]{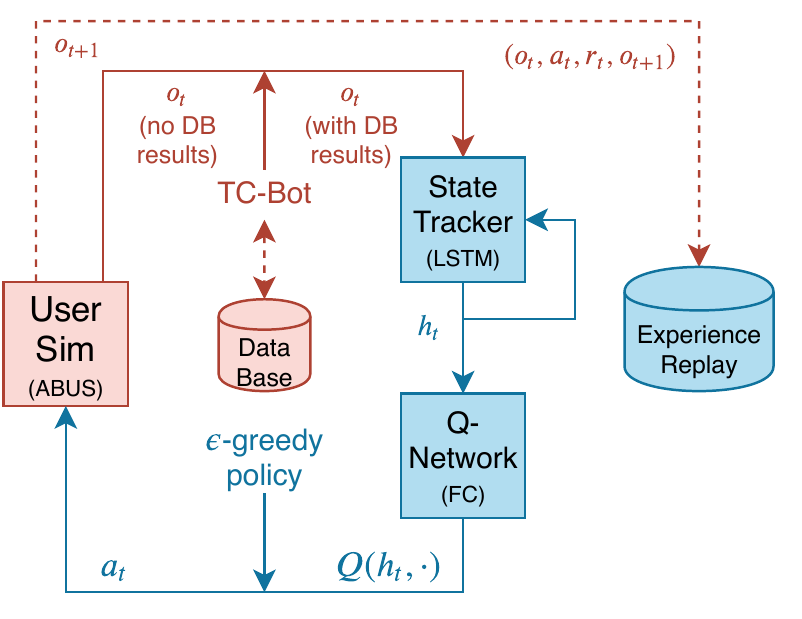}
						\caption{\textbf{Proposed approach overview. The ABUS produces an observation, augmented with TC-Bot knowledge sub-vectors. This observation $o_t$ goes through the state tracker, which yields a belief state $h_t$. The Q-Network predicts a Q-value for each action for the given $h_t$. Then, an $epsilon$-greedy policy is applied to choose an action $a_t$. Finally, a new observation $o_{t+1}$ and a reward $r_t$ are produced, and a transition tuple is stored in the experience replay buffer.}}
						\label{fig:structure}
					\end{figure}
      \end{minipage}
      \hspace{0.05\linewidth}
     \begin{minipage}{0.5\linewidth}
       \begin{figure}[H]
					\includegraphics[width=\textwidth]{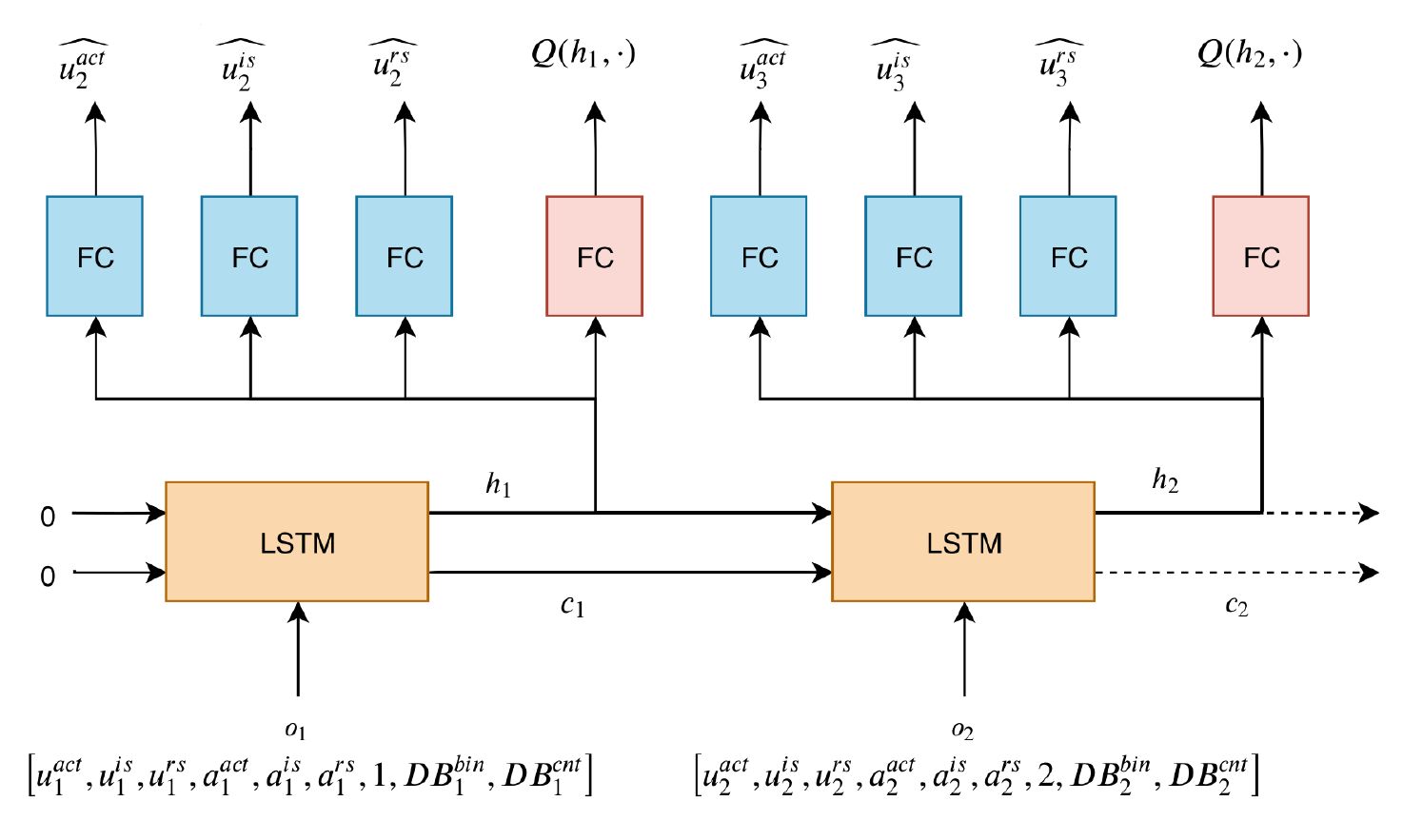}
						\caption{\textbf{Model architectures. DRQN is composed of an LSTM (orange) and a DQN Fully-Connected Q-network (FC, red) layer. The hybrid SL and RL extends DRQN with 3 FC heads (blue) predicting the user behaviour.}}
						\label{fig:models}
			\end{figure}
    \end{minipage}
  \end{minipage}

Our first version was inspired by Deep Recurrent Q-Network (DRQN)
\cite{hausknecht15drqn}, a model that performed quite well in modified
versions of ALE. Unlike in \cite{hausknecht15drqn}, there are no convolutional
layers in our approach, so our model becomes a two-part neural network: a recurrent network
that transforms a sequence of observations $\left[ o_0, ..., o_t \right]$
into a latent vector $h_t$ (belief state), and a fully-connected network on
top to compute Q-values $Q(h_t, \cdot)$ (see \fig \ref{fig:structure}). Moreover, we have used classic
improvements like Experience Replay (ER), a target network, and we explored
the use of ER improvements introduced in \cite{kapturowski18rer} for RRL,
namely stored hidden states and burn-in (unrolling on some early steps of a
sequence without actually computing gradients). We also used domain specific
approaches like Replay Buffer Spiking (RBS) \cite{lipton18bbq}, which partially
fills the ER buffer with a rule-based agent experience in order to introduce
prior knowledge on the rewards by imitation learning.

However, this first version, which separates the learning of ST and agent, did not achieve the performance we were looking
for, so we adopted a joint training procedure as in \cite{li15slrl}. The
authors claim hybrid Supervised Learning (SL) and Reinforcement Learning (RL)
models allow to learn representations of history -- using SL -- which actually
facilitate the learning of the agent -- done using RL --. In fact, by interleaving
Stochastic Gradient Descent (SGD) iterations between SL and RL updates, this
joint training procedure results in a model with richer and general hidden
representation, as it must allow two heads to complete complementary tasks
(choose an action, and predict what will happen). Nevertheless, our
observations are not as simple as in \cite{li15slrl}, so we had to use a
multi-headed network. As can be seen in \fig \ref{fig:models}, the hidden
state of the RNN (which corresponds to the belief state) is passed as input
to multiple fully-connected heads, as well as the DQN head. Each
fully-connected head must predict a different part of the next user-related
part of an observation. The separation of the different user related parts is due to the different meanings and representations they must have. For instance, the $u^{act}$
subvector is one-hot-encoded, meaning there can only be a single activated
dimension, while $u^{is}$ and $u^{rs}$ are binary vectors, meaning multiple
dimensions can be activated. The objective of these subvectors is to be a comparison point during the supervised training to converge towards a robust and meaningful hidden state. 
Thus, during the learning phase the hidden states embeed insights about the future behaviour of the user, useful for the DQN head.
The output of the Q-network head is composed of
the Q-values for each action available for the current state.

As for ER, we tried multiple sampling strategies and sizes of sequences of observations. Maximum success
rate in a single evaluation episode was achieved with the sampling method
described in \cite{kapturowski18rer}: we sample $N$ transition sequences of
length $B + S$, where $B$ is the number of burn-in steps and $S$ is the
number of actual prediction steps. Burn-in steps allow to compute an
intermediate hidden state, so that prediction steps can rely on this vectors
instead of zero vectors. Backpropagation affects prediction steps only.
However, using this method, the model performance decreased after reaching
its peak, and became quite unstable. We conjectured this problem was due to
the shortness of our dialogues as the agent evolves. Thus, we solved this problem, by first
selecting $S$ time steps, and then using all of the preceding time steps as
burn-in steps. Because dialogues are quite short, this does not impact
training time.
\section{Experiments}\label{sec:experiments}

Our experiments aim at examining both the performance of a joint DM training
as well as the interpretability of the learned model.

\subsection{Model Performance}
The curves in \fig \ref{fig:success_rate_wonoise} and \fig \ref{fig:success_rate_noise} show the results in terms of performance, they represent the success rate versus models evolution steps in respectively the noiseless and noisy settings. The green line corresponds to the success of the rule-based agent
that we used for RBS. The red line is just an indicator at 90\% success rate. As explained
before, TC-Bot introduces multiple intent-level and slot-level types of
errors, and the authors explain that intent-level errors do not have a
significant impact on the agent. Thus, we mainly focused on simulation
without noise (\fig \ref{fig:success_rate_wonoise}) and simulation with high
slot-level noise (\fig \ref{fig:success_rate_noise}). In our approach, we used the same
hyper-parameters as the baseline, where possible, and the smallest number of neurons for the
added layers, so that all models have more or less the same number of
parameters (see \tab \ref{tab:Model_parameters}). We also used the same simulation parameters for all our experiments, e.g. number of evaluation episodes to compute success rate, number of training episodes before an evaluation episode, etc (see \tab \ref{tab:Simulation_parameters}).

We have noticed that the most important hyper-parameters are the size of the
recurrent hidden state (belief state) and experience replay. Small hidden states (16, 32) can actually be used to achieve similar results to DQN, in the noiseless setting. However, performance decreases when the hidden state is small and noise is introduced. Thus, we had to increase the number of hidden neurons to 64, which is still rather small.   

In the noiseless setting, the purely RRL model (not presented in the figures) performs slightly worse than the baseline. It converges slower and to a lower success rate. Our hybrid model learns faster than the baseline and has lower variance, at the beginning. However, it achieves the same performance asymptotically and variance increases once the model has achieved its peak performance. Nevertheless, in the noisy setting, our model perform better than the baseline, in terms of learning speed, maximum and asymptotic success rate and dialogue length. These results suggest that a recurrent state tracker leads to a more robust model. In fact, slot-level errors cannot be identified by DQN, while our model can, to some extent. In fact, as our model takes all preceding observations when building the belief state, the resulting state can integrate patterns of a sequence of observations, and thus correct noisy patterns. Moreover, the SL training further enforces this as the model has to learn patterns in user behaviour, as previously mentioned in Section \ref{sec:proposed_approaches}.

Besides, we have studied the contribution of each component of our hybrid model over the overall performance.
In that sense, \fig \ref{fig:success_rate_hidden_state} presents the performance results comparison of our full hybrid model proposed approach, the baseline DQN and the hybrid model based on the TC-Bot current slots instead of the hidden states of our recurrent state tracker. We notice that the hybrid model with \textit{current slots} starts to learn faster than DQN but achieves the worst success rate asymptotically. This confirms that noise robustness of our hybrid model relies mainly on the information yielded by the recurrent state tracker through a richer belief state. Moreover, the results of the \textit{current slots} hybrid model are worse than the DQN baseline while sharing the same belief state.  This let us confirm the positive impact of the joint training procedure between the state tracker and the RL agent.

%

  \begin{minipage}{\linewidth}
      \centering
      \begin{minipage}{0.45\linewidth}
          \begin{figure}[H]
              \includegraphics[width=\linewidth]{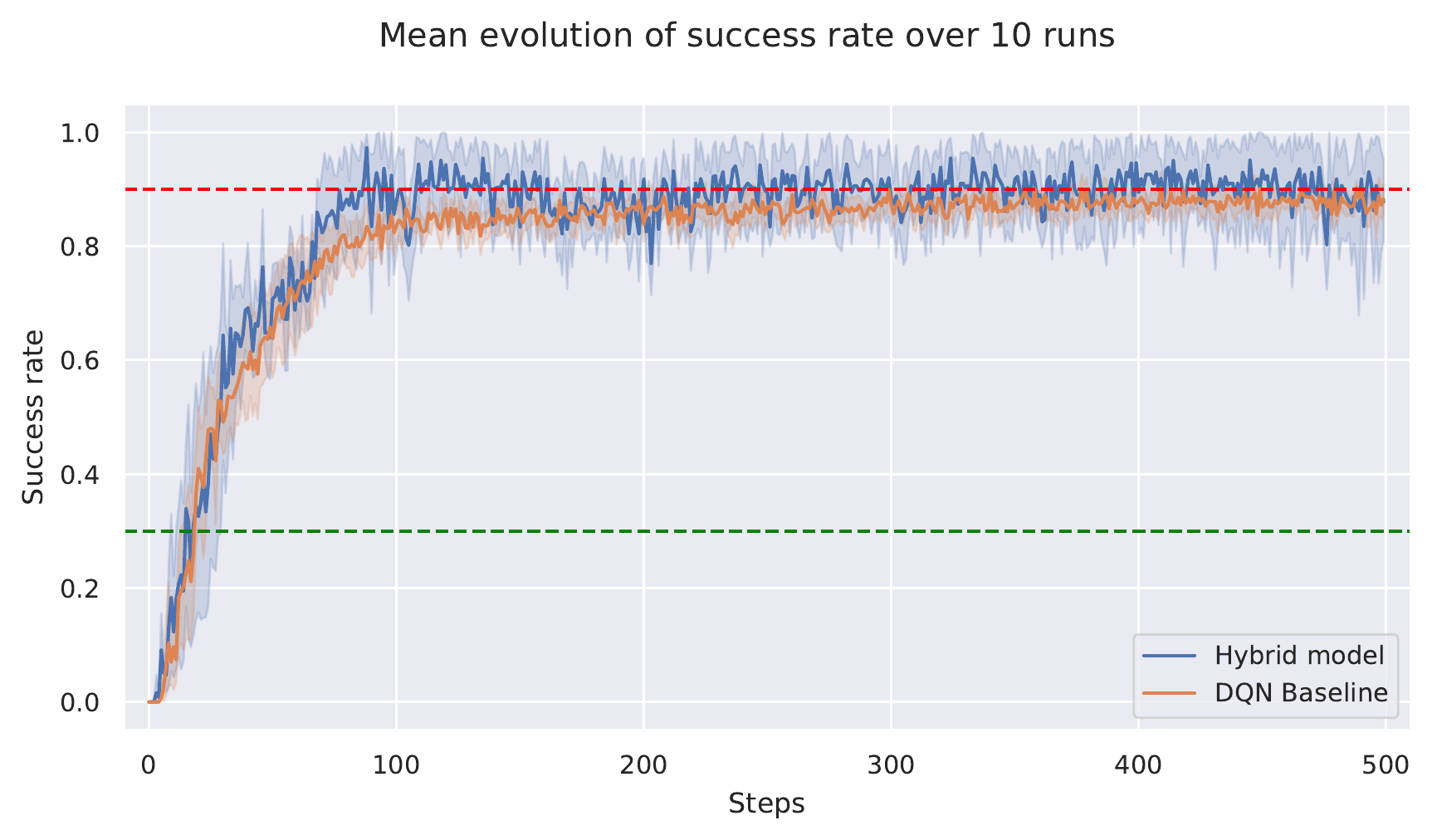}
              \caption{\textbf{Success rate in the noiseless setting. Our hybrid model converges faster than DQN and achieves similar success rate asymptotically, but presents more variance.}}
							\label{fig:success_rate_wonoise}
          \end{figure}
      \end{minipage}
      \hspace{0.05\linewidth}
      \begin{minipage}{0.45\linewidth}
          \begin{figure}[H]
              \includegraphics[width=\linewidth]{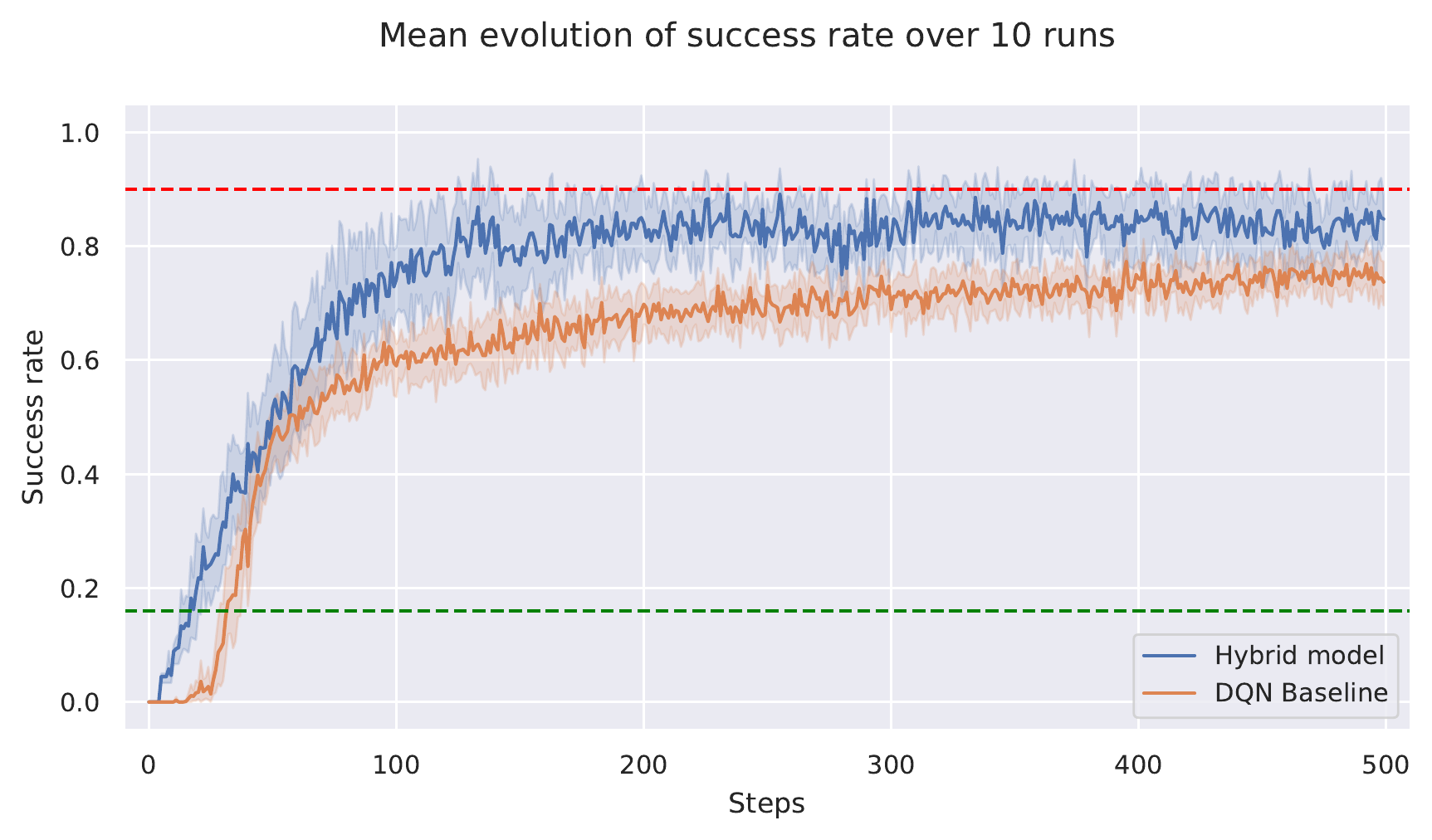}
              \caption{\textbf{Success rate in the noisy setting. Our hybrid model still converges faster, but presents more variance than the baseline as well. The asymptotic success rate is above the baseline by $3\%$.}}
							\label{fig:success_rate_noise}
          \end{figure}
      \end{minipage}
  \end{minipage}

\subsection{Model interpretability}

Explainability has become a highly sought characteristic for ML models. Hence, we examined:
\begin{itemize}
    \item \textbf{Explainability} of the learning process of our model through its evolving policy.
    \item \textbf{Interpretability} of the learned state tracker through intermediate hidden states.
\end{itemize}

\fig \ref{fig:policy_evolution} shows the evolution of the distribution of selected actions. Dialogues illustrating these changes are presented in \tab \ref{tab:policy_evolution}. Policy distribution over time shows three steps in the learning process:
\begin{enumerate}
    \item The agent starts by exploring actions that were not explored by the rule-based agent, as it does not have accurate values for these actions. The distribution shows some peaks on these particular actions.
    \item The policy does not show the peaks, but is quite unstable. In this part, the agent continues to explore, now knowing which actions are truly useless.
    \item The policy stabilizes as the agent learns to ask for the proper information in order to obtain a positive result.
\end{enumerate}

From this, we can see that the state tracker actually produces useful hidden states from the start, as it allows for the agent to choose unexplored strategies. Moreover, as the representation is changing due to RL and SL training, it inherently keeps introducing exploration during the training.

 \begin{minipage}{\linewidth}
      \centering
      \begin{minipage}{0.45\linewidth}
          \begin{figure}[H]
              \includegraphics[width=\linewidth]{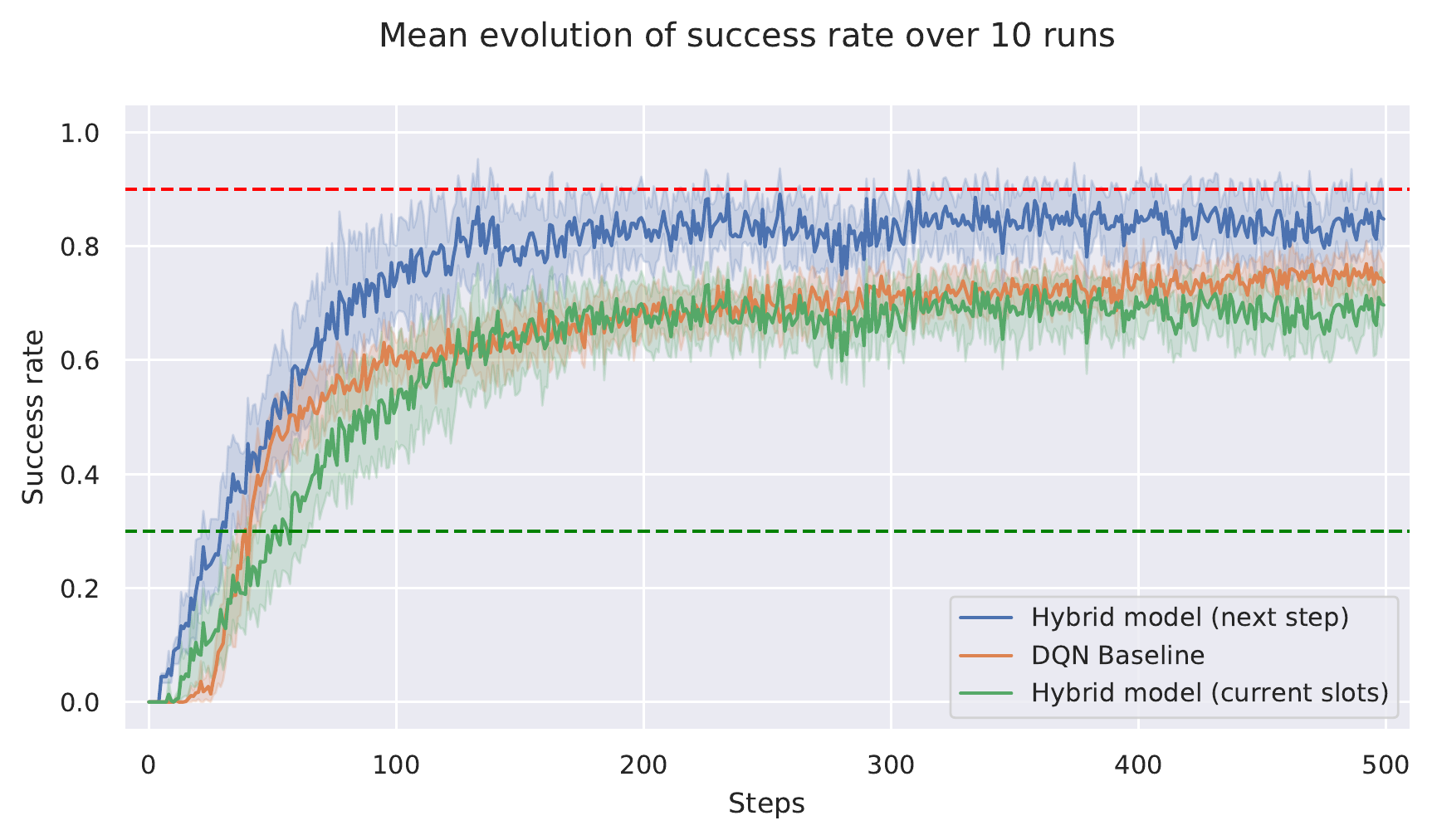}
              \caption{\textbf{Impact of the hidden state on the success rate in the noisy setting. The hybrid model (current slots) starts to learn faster than DQN but achieves the worst success rate asymptotically; this shows both the importance of the hidden state and the joint training procedure between the state tracker and the RL agent in our hybrid model (next step).}}
							\label{fig:success_rate_hidden_state}
          \end{figure}
      \end{minipage}
      \hspace{0.05\linewidth}
      \begin{minipage}{0.45\linewidth}
          \begin{figure}[H]
              \includegraphics[width=\linewidth]{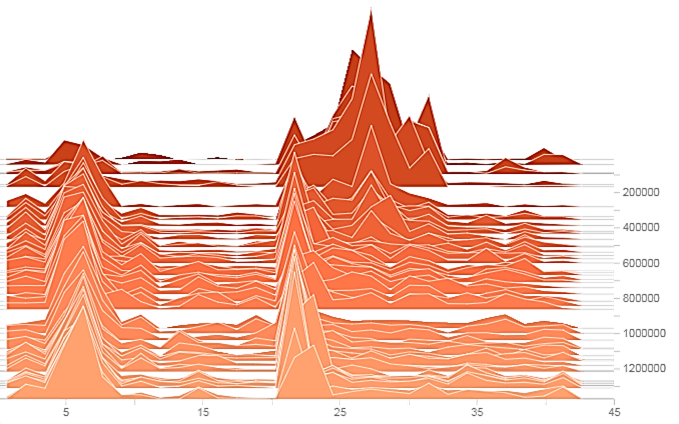}
              \caption{\textbf{Evolution of policy distribution (horizontal and vertical axes) through time (depth axis). At the back, one can see some peaks around the actions 25 through 35 (requesting futile information), but after some training, the policy becomes quite stable (requesting and giving the ticket reference to the user).}}
							\label{fig:policy_evolution}
          \end{figure}
      \end{minipage}
  \end{minipage}

As for hidden state interpretability, \fig \ref{fig:state_tsne} and \fig \ref{fig:state_reg} show respectively a T-SNE embedding visualization using multiple hues and a correlation heatmap between learned hidden states and the \textit{current slots} sub-vectors yielded by TC-Bot. In order to further improve interpretability, we added an $\ell_1$ penalty on the hidden states. Thus, our model produces sparse vectors that can be directly compared to binary representations, such as the \textit{current slots} sub-vector.

The T-SNE visualization allows to see that the hidden states can actually be separated based on dialogue number, time steps and terminal states. This means that our model's representation evolves and stabilizes over time, that it produces consistent representations for the belief state and that it gives insight into when the dialogue might end. It also shows that hidden states cannot easily be separated based on the position of the dialogue in the evaluation episode, which is normal.

The heatmap in \fig \ref{fig:state_reg} between belief states and TC-Bot \textit{current slots} shows that some dimensions better discriminate some classes. To assess the correspondence of certain dimensions to particular boolean vector of TC-Bot current slots vector, we trained multiple random forest classifiers with a One-vs-Rest strategy, with our hidden states as inputs and \textit{current slots} as targets and measured the feature importance of each dimension to each target. This does not allow to find relations between dimensions but rather which dimension best served in classification of a given class of \textit{current slot}.

 \begin{minipage}{\linewidth}
      \centering
      \begin{minipage}{0.45\linewidth}
          \begin{figure}[H]
    \centering
    \begin{subfigure}[b]{0.45\textwidth}
        \includegraphics[width=0.9\textwidth]{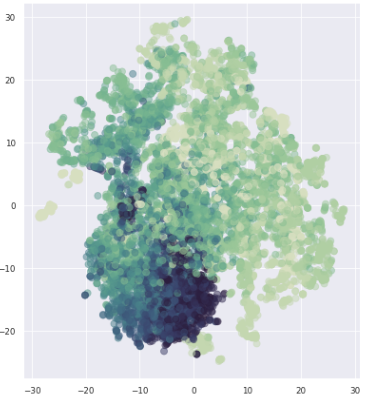}
        \caption{Turn within dialogue. Lighter points are close to 0, darker points are close to 16 (maximum dialogue length).}
        \label{fig:tsne_turn}
    \end{subfigure}
    \hfill
    \begin{subfigure}[b]{0.45\textwidth}
        \includegraphics[width=0.9\textwidth]{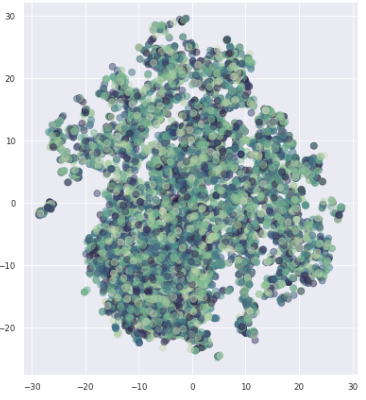}
        \caption{Dialogue within simulation. Lighter points are close to 0, darker points are close to 100 (simulation epoch size).}
        \label{fig:tsne_dialogue}
    \end{subfigure}

    \begin{subfigure}[b]{0.45\textwidth}
        \includegraphics[width=0.9\textwidth]{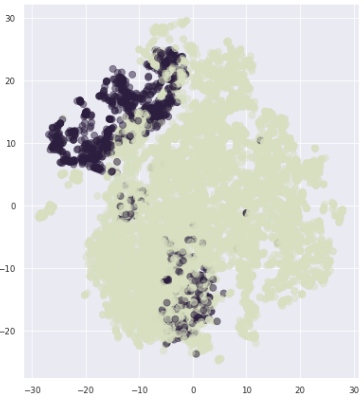}
        \caption{Evolution of terminal steps. Light points are non terminal steps, dark points are terminal steps.}
        \label{fig:tsne_terminal}
    \end{subfigure}
    \caption{\textbf{T-SNE embedding of hidden states according to (a) dialogue length, (b) simulation epoch round, (c) termination step.}}
    \label{fig:state_tsne}
\end{figure}
      \end{minipage}
      \hspace{0.05\linewidth}
      \begin{minipage}{0.45\linewidth}
          \begin{figure}[H]
    \includegraphics[width=0.9\textwidth]{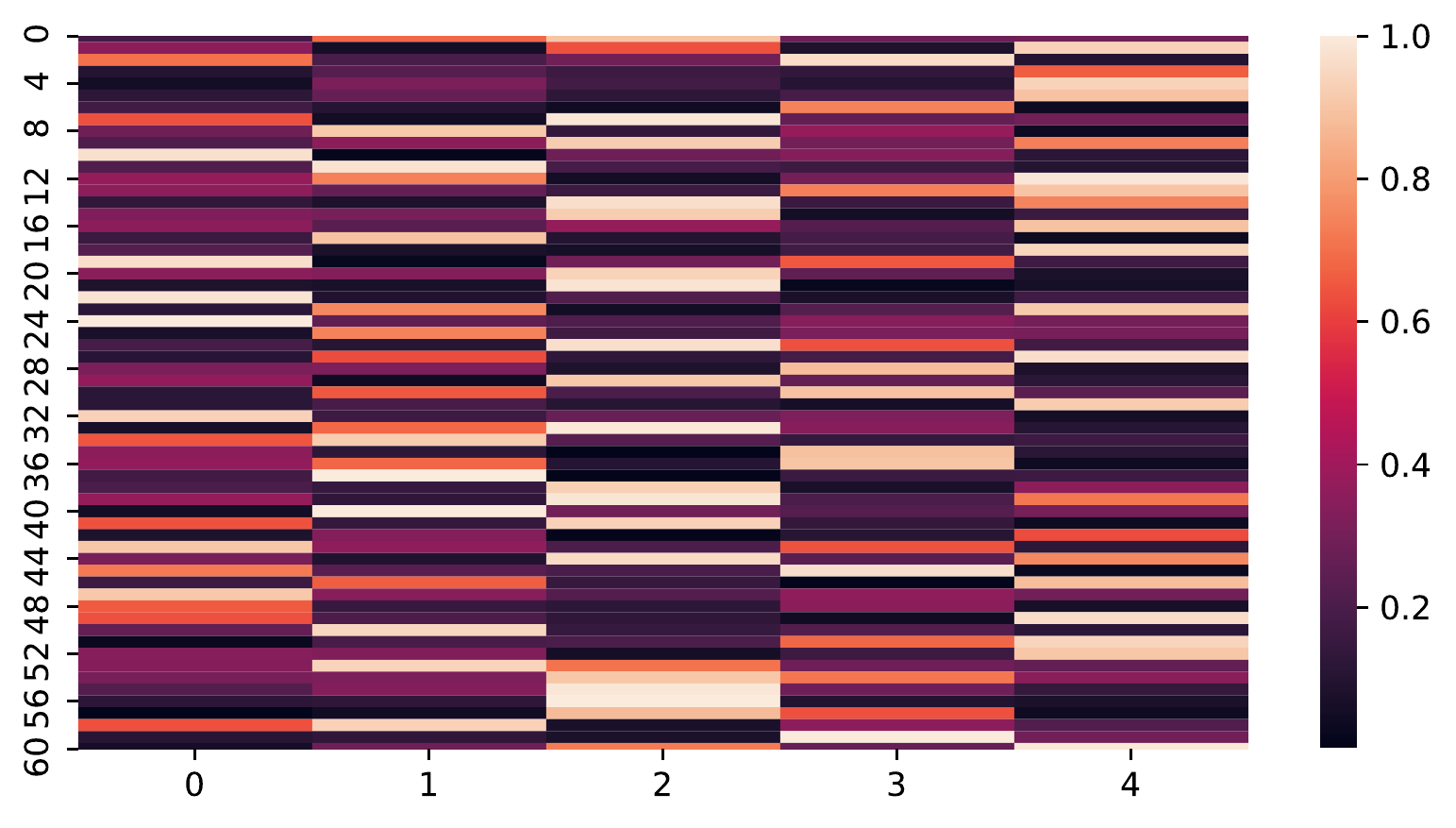}
    \caption{\textbf{Relation between belief states and TC-Bot current slots sub-vector based on feature importance of a One-vs-Rest strategy.}}
    \label{fig:state_reg}
\end{figure}
      \end{minipage}
  \end{minipage}

\section{Related works}\label{sec:related_works}

Led by upgrowing competitions in state tracking, such as the Dialog System
Technology Challenge (DSTC, 2013-2019) and the Amazon Alexa Prize (2018),
novel approaches have been proposed to tackle this task in different ways.
Going from robust hand-crafted rules \cite{wang13bt} and Conditional Random
Fields \cite{lee13discriminative,ren14markov} to more complex neural-based
architectures for response selection \cite{chao19rapnet} and dialogue generation \cite{serban2017multiresolution}.
Most of this work is done within a Spoken Dialogue System (SDS) framework,
which mainly differs from the setting presented in Section
\ref{sec:introduction} in that raw inputs and final outputs are
not text, but go through Speech-To-Text (STT) and Text-To-Speech (TTS)
engines, respectively, so there is additional noise in their observations.

In \cite{henderson13wordbaseddst}, Henderson et al. proposed a belief tracker based on
recurrent neural networks. Their approach takes an Automatic Speech
Recognition (ASR) output of the belief state update, avoiding the use of
complex semantic decoders while still attaining state-of-the-art performance.
Based on this work, Mrkšić et al. proposed a hierarchical procedure to train
a multi-domain RNN dialogue state tracking generalized model \cite{mrksic15mdstrnn}.
The output of their state tracker is a distribution over slot-value pairs.\\

Williams and Zweig \cite{williams16slrl}, as well as Zhao and Eskenazi
\cite{zhao16e2erl}, first introduced recurrent reinforcement learning (RRL) in a
dialogue system. The differences between RRL and previous work are their
objective and the connection between sub-models at training time. In
\cite{henderson13wordbaseddst}, an RNN is used to keep track of the belief state and
outputs a distribution over slot-value pairs. Then, this representation is
used by a policy to choose an action. In \cite{williams16slrl}, an LSTM
single-layer model alternates between supervised learning and policy gradient
\cite{williams92connrl} in order to directly yield a distribution over actions. This approach does not allow to use value-based RL methods. In
\cite{zhao16e2erl}, the authors integrate both of these approaches into a hybrid training procedure with which a
model learns to map sequences of states to a belief state using SL, and
learns an optimal policy using RL. Our approach is
mostly similar to the latter, in that we use a recurrent network for policy
optimization and a hybrid model. However, we use the recurrent layer to
encode a more general history representation, taking as input observations in
the form of semantic frames, instead of constraining the model to do both
slot-filling and action selection with RL as in \cite{zhao16e2erl}.
\section{Conclusion}\label{sec:conclusion}

This paper introduces a new multi-headed hybrid model to perform Dialogue
Management. Starting from the conjecture that a simple state tracker cannot
readily summarize the behaviour leading to a state, but only the available
information, we proposed a hybrid learning approach to learn better belief
states for a Reinforcement Learning agent. Our work shows that active hybrid
learning can help in learning representations containing information about the
possible future states, as seen in the interpretation of the hidden states. This would then be exploited by an RL algorithm to choose the best fitting action as shown in the performance results.

Further work and experimentations could focus on trying different RL algorithms as well as trying to reduce the variance of the proposed model. Moreover, it would be interesting to check if our model is well suited in multi-domain task-oriented dialogues.

\appendix

\section{Appendix}

\begin{table}[H]
 \begin{minipage}{\linewidth}
      \centering
      			\begin{minipage}{0.45\linewidth}
                \footnotesize 
								\begin{tabular}[c]{|c|p{1.5cm}|p{1.5cm}|p{1.5cm}|}
            \hline
            Turn & Dialogue example at $t = 0$          & Dialogue example at $t = M$          & Dialogue example at $t = P$           \\ \hline
            User    & Can I get tickets for \textbf{\textit{Race tomorrow}}? & Can I get tickets for \textbf{\textit{Race tomorrow}}? & Can I get tickets for \textbf{\textit{Race tomorrow}}?  \\ \hline
            Agent    & Do you have any \textbf{\textit{distance constraint}}? & Okay                                 & Which \textbf{\textit{city}} would you like?            \\ \hline
            User    & I do not care.                       & I would like to see it at \textbf{\textit{10:00pm}}.   & I want to watch at \textbf{\textit{Seattle}}.           \\ \hline
            Agent    & Do you have any \textbf{\textit{distance constraint}}? & Which \textbf{\textit{city}} would you like?           & What \textbf{\textit{date}} would you like to watch it? \\ \hline
            User    & I do not care.                       & I want to watch at \textbf{\textit{Seattle}}.          & I want to set it up \textbf{\textit{tomorrow}}.         \\ \hline
            Agent    & What \textbf{\textit{MPAA rating}} would you like?     & Okay                                 & \textbf{\textit{Regal Meridian 16}} is available.       \\ \hline
            User    & I do not care.                       & I want \textbf{\textit{2 tickets}}, please!            & I want to watch at \textbf{\textit{9:10pm}}.            \\ \hline
            \end{tabular}
      \caption{\textbf{Dialogue samples over time. At $t = 0$, the agent chooses the same actions repeatedly; these actions correspond to those that were not explored by the rule-based agent. At $t = M$, the agent has found a policy that works: waiting for the ABUS to give the required information. At $t = P$, the agent has finally found a good policy: it asks questions and even suggests values.}}
    \label{tab:policy_evolution}
      \end{minipage}
			      \hspace{0.05\linewidth}
      \begin{minipage}{0.45\linewidth}
			\footnotesize
              \begin{tabular}{|p{4cm}|p{4cm}|}
    \hline
    Parameter                  & Value                           \\ \hline
    LSTM layer                 & In: 197, Out: 64, Time steps: 3 \\ \hline
    SL rate                    & 0.0001 \\ \hline
    FC layer                   & In: 64, Out: 43                 \\ \hline
    RL rate                    & 0.0005 \\ \hline
    ER size                    & 40 000                          \\ \hline
    Initial $\varepsilon$      & 0.9                             \\ \hline
    Minimal $\varepsilon$      & 0.01                            \\ \hline
    Linear $\varepsilon$ decay & 0.001                           \\ \hline
    Double Learning            & True                            \\ \hline
    Dueling                    & True                            \\ \hline
    Discount factor $\gamma$   & 0.9                             \\ \hline
    Target update frequency    & after each training epoch       \\ \hline
    Polyak averaging $\tau$    & 0.001                           \\ \hline
    Initial learning rate (SL) & $10^{-4}$                       \\ \hline
    Initial learning rate (RL) & $5 \cdot 10^{-4}$               \\ \hline
    Gradient Clipping Range    & {[}-10, 10{]}                   \\ \hline
    Batch size                 & 8                               \\ \hline
    \end{tabular}
    \caption{\textbf{Model parameters for our best runs.}}
		 \label{tab:Model_parameters}

\footnotesize
    \begin{tabular}{|p{4cm}|p{4cm}|}
    \hline
    Parameter                 & Value \\ \hline
    RBS steps                 & 300   \\ \hline
    Evaluation steps          & 100   \\ \hline
    Level 2 error probability & 0.20  \\ \hline
    \end{tabular}
    \caption{\textbf{Simulation parameters for our best runs.}}
		\label{tab:Simulation_parameters}
      \end{minipage}
  \end{minipage}
\end{table}

\newpage
\bibliographystyle{acl}
\bibliography{biblio}

\end{document}